# Short-term wind speed forecasting model based on an attention-gated recurrent neural network and error correction strategy

Haojian Huang[1,*]

haojianhuang927@gmail.com

**Abstract:** The accurate wind speed series forecast is very pivotal to security of grid dispatching and the application of wind power. Nevertheless, on account of their nonlinear and non-stationary nature, their short-term forecast is extremely challenging. Therefore, this dissertation raises one short-term wind speed forecast pattern on the foundation of attention with an improved gated recurrent neural network (AtGRU) and a tactic of error correction. That model uses the AtGRU model as the preliminary predictor and the GRU model as the error corrector. At the beginning, singular spectrum analysis (SSA) is employed in previous wind speed series for lessening the noise. Subsequently, historical wind speed series is going to be used for the predictor training. During this process, the prediction can have certain errors. The sequence of these errors processed by variational modal decomposition (VMD) is used to train the corrector of error. The eventual forecast consequence is just the sum of predictor forecast and error corrector. The proposed SSA-AtGRU-VMD-GRU model outperforms the compared models in three case studies on Woodburn, St. Thomas, and Santa Cruz. It is indicated that the model evidently enhances the correction of the wind speed forecast.

## 1 Introduction

Since the 21st century, clean and renewable energy has been of great significance to mitigating the greenhouse effect and environmental protection. The advancement and application of clean and effective reproducible energy have been the core of studies in the energy field [1]. Since 2020, despite the impact of COVID19, the novel installed capacity of international wind power has reached a novel elevated as ever, driven by large markets such as China and the United States. Towards end of 2020, the world had accumulated an installed 743GW capacity, which represents a 14.3% increase from the previous year. This demonstrates that on account of the advancement of wind power technology and the continual expense reduction, the energy of wind is becoming an increasingly important part of the green energy sector, as global efforts towards an economy with a low carbon continue to gain momentum. The growth of that energy technology is anticipated to offer significant support. Despite the growing number of wind turbines installed worldwide, there are still several countries and regions that do not fully capitalize on their wind energy potential. For instance, Turkey has an estimated wind energy potential of 50,000MW, but its current installed capacity is only approximately 10% of that capacity. Furthermore, various factors such as temperature, air pressure, terrain, altitude, and latitude can affect wind patterns, making it a highly unpredictable and volatile meteorological element. Wind is a complex and challenging variable to forecast, given its non-linear, non-stationary nature, and random characteristics [2]. The power grid's utilization of wind power and its dispatching safety face challenges at different times and in diverse terrains. These challenges can compromise the power system's stability and lead to increased costs in the power market. To ensure the safe and efficient use of the mentioned energy, it`s crucial to make precision predictions of wind speed and its characteristics. This can facilitate optimal utilization of installed wind power capacity, increase wind energy's utilization rate, and enable the safe dispatch in one large-scale network, thus enhancing this power's security. As such, the development of precise and timely short-run multi-step prediction of wind speed has become an urgent technology.

Currently, the dominating wind speed prediction approaches is able to be split into physical approaches, artificial intelligence approaches, spatial correlation approaches, statistical approaches, along with combined model approaches in terms of the principle [3]. Specifically, the physical approaches demand to rest with the detailed forecast concerning weather, environmental factors information, such as temperature, humidity, pressure, and other data for prediction, which requires a tremendous and complex calculation [4]; Statistical methods are to establish a statistical, mathematical model according to the historical wind speed series for extrapolation [5]. The spatial correlation methods can expounded the spatial correlation among diverse observation sites and the relationship among all sites within a particular area to achieve wind speed prediction in a specific spatial area [6]. However, the drawbacks of these methods are also obvious. Models including Numerical Weather Prediction (NWF) [7], Model Output Statistics (MOS) [8] and Eta Models [9], High-Resolution Models (HRM) [10] often combine multiple physical factors to provide a satisfactory prediction. Nevertheless, these models are computationally expensive and complex, not easy to apply [11]. By contrast, statistical models can fully exploit the hidden information of historical data through mathematical patterns, such as autoregressive (AR) [12], autoregressive moving average (ARMA) [13], and autoregressive integrated moving average (ARIMA) [14]. For example, Q. Xu et al. adopted the ARIMA pattern and wavelet transform to forecast the wind speed series [15]. What is more, the exponential smoothing also makes a difference [16]. Although short-run wind speed predicting approaches, such as those on the foundation of prior assumptions about wind speed data distribution, have demonstrated high predictive accuracy, their effectiveness is limited because they do not account for its nonlinear and non-stationary essence [17]. Additionally, while spatial correlation methods can reveal wind speed distribution characteristics in an area, implementing such models can be challenging since they require measurements of wind speed from multiple spatially correlated locations and have stringent requirements for measurement error and time delay [18].

*Corresponding author

With the aim of enhancing the precision and feasibility of wind velocity predictions, artificial intelligence methods have become a technology that scholars have paid extensive attention to in recent years [19]. Compared with the first three types of methods, they are able to abstract the main nonlinear features from historical wind speed series functioning as a basis for prediction, and achieve better consequences than first three categories of approaches. When it comes to artificial intelligence methods, in terms of[20], the prevalent techniques primarily rely on artificial neural networks (ANN) and support vector machines (SVM). Specifically, ANN-based methods, including back-propagation neural network (BPNN) [21], extreme learning machine (ELM) [22], generalized recurrent neural network (GRNN) [23], recurrent neural network (RNN) [24], and Elman neural network (ENN) [25], which are broadly applied to the domain of the wind speed forecast, financial forecast and power load forecast, have already achieved good forecasting results. However, the prediction accuracy of a single agent is subject to limitations especially when it deals with complicated sequences, and it often takes a long time to obtain good enough prediction results. To be more specific, BPNN is prone to getting trapped in local optima, ELM is susceptible to the randomly generated initial weights and bias, and RNN tends to suffer from the problems of gradient vanishing and exploding when the sequence is long and the network structure is too deep. To tackle these issues, some researchers have adopted various intelligent optimization algorithms or heuristic mathematical models to optimize artificial intelligence approaches. Though approaches such as genetic algorithm (GA) [26] and optimization of the particle swarm (PSO) [27] can optimize network parameters and enhance prediction accuracy, artificial neural networks still have a simple structure and shallow layers, which limits their ability to extract abstract sequence features. Therefore, the aforementioned challenges have not been fully addressed. Currently, the deep neural networks has already been employed in the related forecast for the purpose of addressing issues, too. By comparison with the shallow artificial neural network model, the deep neural network pattern own one more complicated hidden layer structure, which can better extract the abstract features and latent patterns in the sequences. The basis of deep learning methods is the RNN. Although RNN has fulfilled some consequences in handing the time series issues, the issues of gradient explosion and disappearance still exist. The essence of these problems is related to the mechanism that its training relies on long-term iteration and backpropagation. The long short term memory (LSTM) adds a memory unit to the network structure, which better overcomes the problem of gradient disappearance. At the same time, it uses the gradient clipping mechanism to truncate when the gradient exceeds a certain threshold, effectively solving the problem of gradient explosion [28]. However, LSTM owns a prodigious quantity of the parameters, which brings about the difficulty of convergence of training in the effective time. The gated recurrent unit (GRU) simplifies the structure of LSTM, and takes the place of the input gate with an "update gate" as well as the forgetting gate, which makes the network construct simple, reduces parameters, and significantly ameliorate the convergence speed of network training. Even though the historical wind speed series has a small sample size, GRU can still maintain excellent prediction performance [29] .

Nonetheless, wind patterns vary between different areas, which means that a singular model forecast cannot ensure that it is the most optimal for every location. Simultaneously, relying on a solitary artificial intelligence approach could be challenging to avoid being stuck in a suboptimal solution, while having a slow rate of convergence.Studies have shown that artificial neural network methods perform well in mixed models [30]. Moreover, as a mixture of statistical methods, traditional neural networks, along with further learning architectures, combined prediction models could greatly learn nonlinear and linear features, significantly improving predictions' accuracy. At present, the workflow of most combined prediction models is split into three sections: figures preprocessing, construction of suitable prediction models, and optimization of model parameters [31]. Data preprocessing can reduce non-stationarity to extract the main features of sequence and improve the forecast precision of the pattern. Currently, there are many preprocessing strategies on the foundation of the sequence decomposition, including wavelet transform (WT) [32], empirical mode decomposition (EMD) [33], ensemble empirical mode decomposition (EEMD) [34], complete ensemble empirical mode decomposition with adaptive noise (CEEMDAN) [35], singular spectrum analysis (SSA) [36], and variational mode decomposition (VMD) [37]. It`s difficult for the EMD to fulfill the satisfactory consequences in dissecting nonlinear and non-stationary sequences among the above strategies because of its decomposition effect is easily affected by complex modal mixing problems. To tackle this issue, EEMD was developed to incorporate Gaussian white noise into EMD. Nevertheless, EEMD often generates residual noise that can produce imaginary modes in the reconstructed signal, leading to suboptimal performance. CEEMDAN introduced specific noise into each decomposition stage to prevent the generation of imaginary modes in the reconstructed sequence. However, CEEMDAN is prone to poor parameter robustness, which can lead to unsatisfactory prediction results. By contrast, VMD ensures that all wind speed components remain orthogonal throughout the decomposition process, which significantly reduces the likelihood of modal aliasing [38].

Due to the consistent patterns in wind power generation, the attention module is used to assign varying weights to historical data. This approach helps isolate significant trends from less relevant information in the wind speed series, enhancing the analysis of key characteristics for more accurate forecasting. Therefore, recently, attention module is employed in the area of the wind speed series forecast[39-45]. It is evident that numerous composite models currently rely on the optimization of each pertinent parameter in the prediction model, or attempt to decide the best combination of weights for multiple patterns [46]. Meanwhile, forecast consequences of every sub-model are added to get a more accurate short-term prediction [47-48]. However, models based on this type of strategy usually require much preparatory work to achieve good prediction results. In contrast, in [49], a multi-step forecast pattern which combines researches concerning weather and prediction simulation and strategy of the error correction is proposed. This pattern obviously enhances the accuracy of multi-step preliminary prediction of wind speed series on the foundation of the strategy of error correction. Its complexity is low. Recurrent neural networks and reinforcement learning have been applied in [50], respectively, based on the error correction strategy. Unfortunately, the error correction strategy is still not developed great enough in wind speed series forecast.

Therefore, this essay raised one corrector of errors on the foundation of the idea of error correction, and combines the attention to propose a combined short-run wind speed forecast pattern on the foundation of the attention gated recurrent neural network (AtGRU) and the strategy of error correction. The model is made up of the GRU pattern as the error corrector and the AtGRU pattern as the preliminary predictor. First of all, considering the potential noise components in the series of wind speed, SSA is employed with a view to denoising the source wind speeds. Next, the denoised wind speed data is preliminarily predicted using AtGRU. Next, errors are able to be figured out on the foundation of the distinction between existing preliminary predictions and data of historical wind speed series. The error sequence composed of these errors reflects the error distribution characteristics of preliminary prediction. Processed by VMD, the error sequence is split into an array of sub-modal error sequences with limited bandwidth and different characteristics and frequency bands, reducing its nonlinearity and obtaining its main features for the training of the error corrector. Then, the GRU is used to predict each sub-model sequence attained through decomposing, and the prediction is superimposed to attain an error correction value. Finally, the prediction will be superimposed through forecast of that predictor and error corrector. The model's validity is verified on the data-set of three wind farm sites in Woodburn, St. Thomas Island and Santa Cruz Island, respectively.

According to the above work, the dominating contributions of my essay could be summarized as follows:

(i). We introduce a model combining GRUs and an attention module for wind speed forecasting. This model self-adaptively learns critical patterns from historical wind speed data, allowing it to pinpoint essential periods that significantly impact future predictions. By intelligently adjusting to changes in wind dynamics, the model consistently enhances forecast accuracy, proving highly effective in real-world applications.

(ii). Our SSA-AtGRU-VMD-GRU model introduces an advanced error correction mechanism, combining SSA and VMD with GRUs. This design features a preliminary predictor and an error corrector, which collaboratively refine forecasts by correcting deviations. The integration of these elements significantly improves the precision and reliability of wind speed predictions.

(iii). Comprehensive experiments using historical wind speed data validate our model's effectiveness. Results show superior accuracy and robustness over traditional methods, confirming the model's ability to efficiently handle complex wind dynamics and enhance energy system reliability.

In **section 2**, a series of necessary theoretical backgrounds related to the proposed model will be introduced; Inside **section 3**, the entire workflow is exhibited; Inside **section 4**, these consequences of comparative experiments are analyzed one by one; **Section 5** focuses on the advantages of the pattern; **Section 6** sums up the whole article.

## 2 Theoretical basis of the proposed model

### *2.1 Singular spectrum analysis*

SSA is used to decompose original sequences into interpretable subsequences.It is widely applied in biology, engineering, climatology, and economics. The process could be concluded as follows:

**Step 1**(Embedding): the initial time series, $c(c_1, c_2, \cdots, c_N)$ is converted into sequence $z(z_1, z_2, \cdots, z_K)$ through Equation (1).

$$c(c_1, c_2, \cdots, c_N) \to z(z_1, z_2, \cdots, z_K) \tag{1}$$

The dimension embedded in SSA is expressed by $L$. For $K = N - L + 1$, $L$ lag vector can be defined as: $z_i(c_i, c_{i+1}, \cdots, c_{i+L-1})^T \in R^L$, $L \in [2, N]$.And $z = (z_1, z_2, \cdots, z_K)$ is expressed as the trajectory matrix in Equation (2).

$$z = (z_1, z_2, \cdots, z_K) = \begin{pmatrix} c_1 & c_2 & \cdots & c_K \\ c_2 & c_3 & \cdots & c_{K+1} \\ \cdots & \cdots & \cdots & \cdots \\ c_L & c_{L+1} & \cdots & c_N \end{pmatrix} \tag{2}$$

**Step 2** (Singular value decomposition): For the covariance matrix $s = z \cdot z^T$, the singular value decomposition method is used to generate $L$ eigenvalues ( $\lambda_1, \lambda_2, \cdots, \lambda_L$ ) and eigenvectors( $u_1, u_2, \cdots, u_L$ ). If $t = \max(i, such\ that \lambda_i > 0)$ and $v_i = z^T u_i \sqrt{\lambda_i}$ $(i = 1,2, \cdots, t)$, SVD (singular value decomposition) of the trajectory matrix could be expressed to be:

$$z = z_1 + z_2 + \cdots + z_t \tag{3}$$

in which $z_i = \sqrt{\lambda_i} u_i v_i$, with a rank of 1, and its main component $v_1, v_2, \cdots, v_t$. Besides, $\sqrt{\lambda_i} u_i v_i$ is the eigensolution ofSVD of $z$.

**Step 3** (Grouping): The range is decomposed into subsets $a_1, a_2, \cdots, a_m$; and there is no connection among them. The resulting matrix $y_a$ of $a = (n_1, n_2, \cdots, n_p)$ can be expressed as$y_a = y_{a_1} + y_{a_2} + \cdots + y_{a_p}$,and the trajectory matrix is decomposed into $y = y_{a_1} + y_{a_2} + \cdots + y_{a_m}$.

**Step 4** (Diagonal averaging): The primary goal of this process is just to convert per resulting matrix $y_a$into a new series of length $N$. Assuming $d$ is a matrix of $L \times K$,where $L^* = \min(L, K)$ and $K^* = \max(L, K)$. If $L < K$ ,then $y^*_{ij} = y_{ij}$ ;otherwise $y^*_{ij} = y_{ji}$. Then transform $y$ to a sequence $(r_1, r_2, \cdots, r_L)$according to the following equation:

$$r_k = \begin{cases} \dfrac{1}{k+1}\sum_{q=1}^{k+1} y^*_{q,k-q+1}, 1 \le k \le L^* \\ \dfrac{1}{L^*}\sum_{q=1}^{L^*} y^*_{q,k-q+1}, L^* \le k \le K^* \\ \dfrac{1}{N-K+1}\sum_{q=1}^{N-K^*+1} y^*_{q,k-q+1}, K^* \le k \le N \end{cases} \tag{4}$$

As a result, the reconstructed series is able to be obtained by using the first $p$ principal component.

*2.2 Phase space reconstruction*

According to the embedding theorem, for the time series data with chaotic properties, a phase space that is the same as that of the dynamic system in the topological sense, can be reconstructed from the time series. It is susceptible to initial conditions, so the phase space reconstruction approach is often adopted for short-run forecast. The key to reconstructing phase space using the delay coordinate method is the selection of embedding dimension and delay time.

i.Delay time

Define two-time series $S$ and $Q$, as shown in the equation below, where $Q$ and $S$ are obtained after a delay of $\tau$

$$(S,Q) = (x(i), x(i+\tau)), 1 \le i \le (n-\tau) \tag{5}$$

The mutual information between two-time series is a function of delay time $\tau$. The size of mutual information$I(\tau)$ represents the certainty of time series when the time series$S$is known. It can be expressed as follows:

$$I(\tau) = I(S,Q) = H(S) + H(Q) - H(S,Q) \tag{6}$$

Where the first minimum of $I(\tau)$ represents the maximum possibility of uncorrelation between two-time series, so the first minimum of $I(\tau)$ is used as the final delay time;$H(S)$,$H(Q)$ and $H(S, Q)$ are the information entropy and joint information entropy of their respective time series, as shown in Equation (7) - (9)

$$H(S) = \sum_{i}^{m} P(x_i)\log_2 P(x_i) \tag{7}$$

$$H(Q) = \sum_{j=1}^{n} P(x_j+\tau)\log_2 P(x_i+\tau) \tag{8}$$

$$H(S,Q) = \sum_{i=1}^{m}\sum_{j=1}^{n} P(x_i, x_i+\tau)\log_2 P(x_i, x_i+\tau) \tag{9}$$

Where $P$ is the corresponding probability.

ii.Embedding dimension

Based on the time series $S$ and delay time of data $\tau$, phase space can be constructed as $y_i(d)$ shown in Equation (10).

$$y_i(d) = x(i), \ldots, x(i+(d-1)\tau), i = 1,2,3\ldots, n-(d-1)\tau \tag{10}$$

in which $d$ is the embedded dimension.

The so-called false adjacent points refer to pairs of non-adjacent points in a high-dimensional phase space that appear as adjacent points when projected onto a one-dimensional space. In a d-dimensional phase space, each vector $y_i(d)$ has a nearest point. Between the two points, the distance is able to be calculated with either Euclidean distance or the maximum norm, which is adopted in this study. The variables $a(i,d)$, $E(d)$, $E_1(d)$ are defined as follows:

$$a(i,d) = \frac{\| y_i(d+1) - y_{n(i,d)}(d+1) \|_\infty}{\| y_i(d) - y_{n(i,d)}(d) \|_\infty} \tag{11}$$

$$E(d) = (n - d\tau)^{-1} \sum_{i=1}^{n-d\tau} a(i,d) \tag{12}$$

$$\Delta E(d) = \frac{E(d+1)}{E(d)} \tag{13}$$

Where $a(i,d)$ aims to make the corresponding vector closest to the vector $y_i(d)$ under the maximum model number; $E(d)$ takes the average value of $a(i,d)$ within dimension $d$ and $\Delta E(d)$ is the change degree of the variable when the dimension changes from $d$ to $d+1$. Through iterative calculation, when the dimension $d$ is greater than a certain value $d_c$ and $\Delta E(d)$ no longer changes, $d_c+1$ is selected as the embedded dimension of the time series.

iii.Local mean prediction

The reconstruction of the phase space is carried out based on the chosen embedding dimension and delay time, and the $i^{th}$ vector $y_i$ is generated according to Equation (10). Consequently, the system model is established in d-dimensional Euclidean space, which is presented as follows:

$$y_{i+1} = F(y_i) \tag{14}$$

Where $F$ is a continuous function. Let $N = n - (d-1)\tau$, in terms of the properties of continuous functions, if $y_n$ and $y_h$ are very close, $x(n+1)$ can be used as the approximation of $x(N+1)$. Taking the maximum number $\|\cdot\|_\infty$ of models as the measurement standard, the $k$ nearest vector to $y_N$ is selected from the phase space, and the predicted value $x(n+1)$ of the next time in the time series can be obtained according to the local average prediction idea as follows:

$$x(n+1) \approx k^{-1} \sum_{h=1}^{k} x(t_h + 1 + (d-1)\tau) \tag{15}$$

Where $t_h$ is the h^th^ nearest vector.

## *2.3 Gated recurrent unit network*

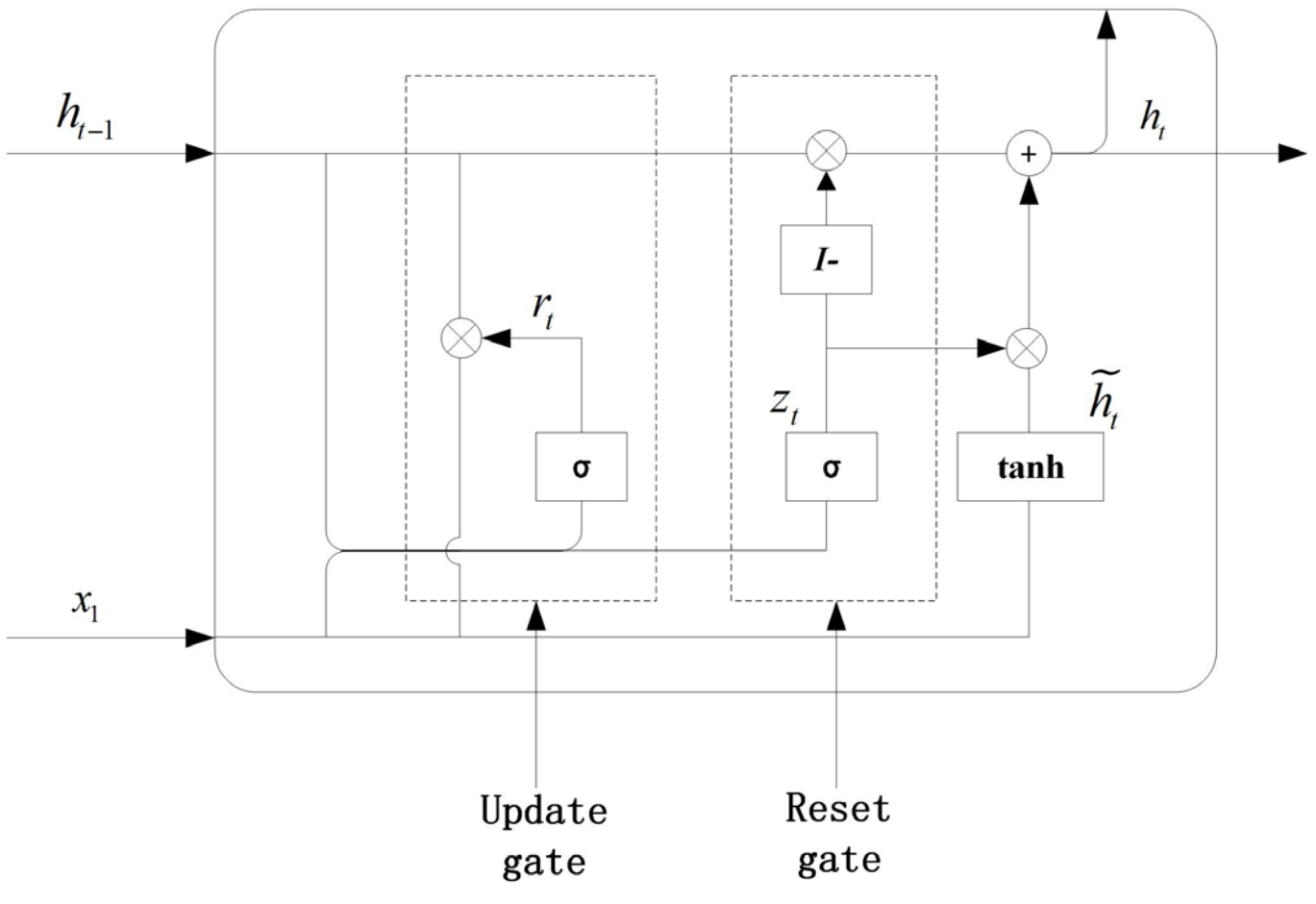

**Fig.1.** GRU neural network structure

$$\begin{cases} r_t = \sigma(W_r \cdot [h_{t-1}, x_t]) \\ z_t = \sigma(W_z \cdot [h_{t-1}, x_t]) \\ \tilde{h} = tanh(W_{\tilde{h}} \cdot [r_t \times h_{t-1}, x_t]) \\ h_t = (I - z_t) \times h_{t-1} + z_t \times \tilde{h}_t \\ y_t = \sigma(W_o \cdot h_t) \end{cases} \quad (16)$$

The network of GRU is an enhanced version of LSTM network that improves gate functions of both long-run and short-run memory units. It blends the gate of input and forget the gate into one single update gate and compounds the hidden and neuron states. This efficiently addresses the issue of "gradient disappearance" inside recurrent neural networks, lessens the quantity of parameters inside the units of memory, and speeds up the time of training. The network's foundation construction is described in **Fig. 1.** Besides, its mathematical representation is described in Equation (16).

In **Fig. 1** as well as Equation (16), $x_t$, $h_{t-1}$, $h_t$, $r_t$, $z_t$, $\tilde{h}_t$, $y_t$refer to the vector of input, the state variable of memory for previous time step, the state variable of memory for current time step, the state of update gate, the state of reset gate, the state of present candidate set, along with output vector of present time step, respectively. $W_r$, $W_z$, $W_{\tilde{h}}$, $W_o$ are just weight parameters of which the link matrix created by $x_t$ and$h_{t-1}$multiplied by the gate of update, the gate of reset, the set of candidate, and the vector of output, separately; $I$ stands for the matrix of identity; $\sigma$denotes the function of activation; $tanh$ is the function of tangent.$W_r$, $W_z$, $W_{\tilde{h}}$, $W_o$ are weight parameters that correspond to the connection matrix formed by input vector $x_t$ and state memory variable $h_{t-1}$ and are multiplied by the gate of update, the gate of reset, the set of candidate, and the vector of output, separately. The symbol $I$ presents the matrix of identity, σ presents the function of activation, and tanh presents the hyperbolic function of tangent.

Central components of the GRU network are the update and reset gates. The variable of input $x_t$ and the split matrix of the state memory variable $h_{t-1}$ from the previous time step are passed through the update gate by means of a nonlinear transformation with sigmoid, which decides how much of the prior state variable need to be put into the present state. The gate of reset adjusts the quantity of data written into the set of candidate from the prior time. The data from the prior time is stored as $h_{t-1}$ times $I$ minus $z_t$, while the current time's information is stored as $\tilde{h}_t$ times$z_t$. Their sum is the output of the current time step.

### *2.4 Attention module*

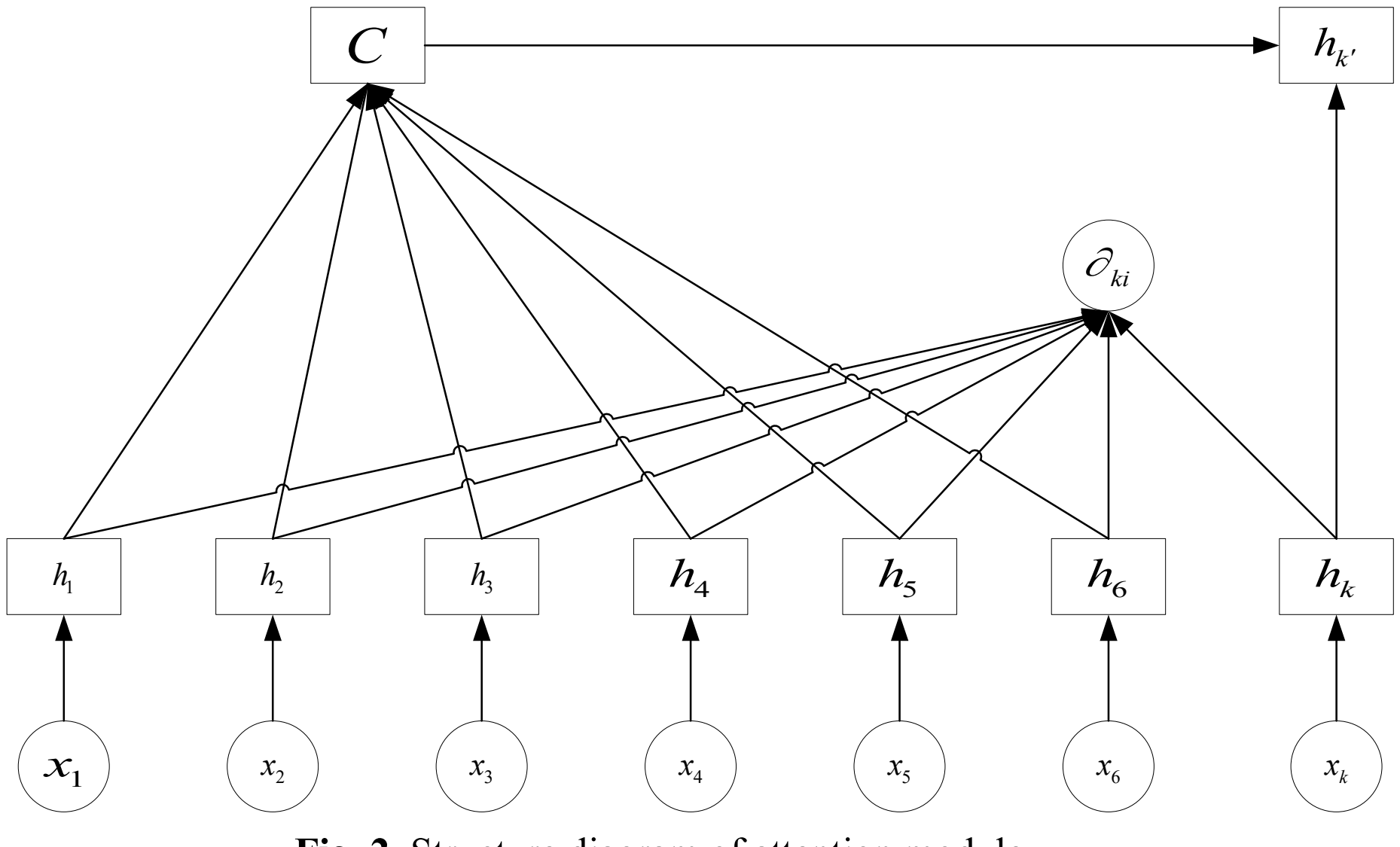

**Fig. 2.** Structure diagram of attention module

The essence of attention is to imitate the distribution of human brain attention. People will allocate extra attention to the information received by the human brain. They will increase their attention where they are interested and ignore their information where they are not interested. In this way, more important information can be obtained with less brainpower. The computer is also designed to imitate the human brain. As the technology of artificial intelligence advances speedily, the computer can process more and more data and information. However, some information in a large amount of data has little effect on data analysis. Some data can even interfere with data mining and judgment. Thus, attention was born. Strictly speaking, it is not an implementation of a model, so the attention has different performance and application methods.The structure of the basic attention model is shown in**Fig.2**.

Where$x_1, x_2, \cdots, x_k$ is the input sequence;$h_1, h_2, \cdots, h_k$is the state value of the hidden layer associated with the input sequencing; $\partial_{ki}$ is just the attention weight value of the hidden layer of the historical input information to the present input information; $C$ is a weighting of the entire hidden state; $h_{k'}$ is the hidden layer state value of last output node.

The input sequence is denoted as $x_1, x_2, \cdots, x_k$ and the corresponding hidden layer state values are represented as$h_1, h_2, \cdots, h_k$. The attention weight value $\partial_{ki}$ indicates the influence of the historical input data on the present input information for the hidden layer. The entire hidden state is weighted by $C$, and $h_{k'}$ refers to the hidden layer state value of the node of the last output.

The attention model is as follows:

$$\partial_{ki} = \frac{\exp(S_{ki})}{\sum_{j=i}^{T} \exp(S_{kj})} \tag{17}$$

$$S_{kj} = v \tanh(W h_k + U h_i + b) \tag{18}$$

$$C = \sum_{i=1}^{T_t} \partial_{ki} h_i \tag{19}$$

$$h_{k'} = H(C, h_k, x_k) \tag{20}$$

Where $S_{ki}$ is the energy value of the hidden layer $h_i$ at the moment $i$, $T_x$ is just the length of input information, $v$ is the input value, $U$ and $W$ is the weight matrix.

### *2.5 Variational mode decomposition*

Konstantin Dragomiretskiy et al. proposed the variational modal decomposition theory in 2014, whose theoretical framework is adaptive to handle the optimal solution of limited variational models. Find $K$ modal functions $u_k(t)\,(k \le K)$ through continuous iteration to determine the central bandwidth and frequency of each Intrinsic Mode Functions (IMFs) component, realize the signal frequency domain division and effective separation of each IMF component, and minimize the sum of all IMFs estimated bandwidth. The specific decomposition process of the VMD algorithm is shown in Equation (21) - (26) :

Define intrinsic modal functions $u_k(t)$, that is

$$u_k(t) = A_k(t) \cos(\varphi_k(t)) \tag{21}$$

Where the signal $\varphi_k$ meets the requirement of non-monotone decreasing, that is $\varphi_k(t)' \ge 0$. And $A_k(t)$ is instantaneous and satisfies $A_k(t) \ge 0$. What's more, the phase values $\varphi_k(t)$ change faster than $A_k(t)$ and the sum instantaneous frequency$\omega_k(t) = \varphi_k(t)'$. The specific solution steps are as follows:

(1)The modal function is expressed as$\{u_k\} = \{u_1, u_2, \cdots, u_k\}$and the new analytic signals are generated by Hilbert Transform;

$$\left[\delta(\mathrm{t}) + \frac{j}{\pi \mathrm{t}}\right] u_k(t) \tag{22}$$

Where$\delta(\mathrm{t})$is the impulse function.

(2)Then estimate the center frequency $\omega_k$, the set of central frequencies is expressed as $\{\omega_k\} = \{\omega_1, \omega_2, \cdots, \omega_k\}$;

(3) The variational constraint model is:

$$\min_{\{u_k\},\{w_k\}}\left\{\sum_{k=1}^{K}\parallel \partial_t\left\{\left[\left(\delta(t)+\frac{j}{\pi t}\right]^* u_k(t)\right\}e^{-j\omega_k t}\parallel_2^2\right\},\ \text{s.t.}\ \sum_{k=1}^{K}u_k=x(t) \tag{23}$$

Where " * " is the symbol of convolution operation.

The multiplication operator alternate direction method is used to update alternately $u_k^{n+1}$, ${\omega_k}^{n+1}$ and $\lambda^{n+1}$ toseek "saddle points" of the extended Lagrange expression. And the modal component $u_k^{n+1}$ is expressed as:

$$u_k^{n+1}=\arg\min_{u_k\in X}\left\{\alpha\parallel \partial_t\left\{\left[\left(\delta(t)+\frac{j}{\pi t}\right]^* u_k(t)\right\}e^{j\omega_k t}\parallel+\parallel f(t)\sum_i u_i(t)+\frac{\lambda(t)}{2}\parallel_2^2\right\} \tag{24}$$

Where $\alpha$is the penalty factor, and $\lambda$is the Lagrange multiplier.

The expression of central frequency ${\omega_k}^{n+1}$ is:

$$\omega_k^{n+1}(\omega)=\frac{\int_0^\infty \omega|\hat{u}_k(\omega)|^2 d\omega}{\int_0^\infty |\hat{u}_k(\omega)|^2 d\omega} \tag{25}$$

The suspensive condition of iteration are as follows:

$$\frac{\sum_{k=1}^{K}\left\|\hat{u}_k^{n+1}-\hat{u}_k^n\right\|_2^2}{\left\|\hat{u}_k^n\right\|_2^2}-e<0 \tag{26}$$

According to the above derivation, the calculation steps of the VMD algorithm are as follows:

**Step1**: Initialization $\{\hat{u}_k^1\}$, $\{\hat{\omega}_k^1\}$,$\{\hat{\lambda}^1\}$ and $\mathrm{n}=0$;

**Step2**: Execution cycle $\mathrm{n}=\mathrm{n}+1$;

**Step3**: Update $\omega_k$ and $u_k$

**Step4**: Update $\lambda$: $\hat{\lambda}^{n+1}(\omega)\leftarrow\hat{\lambda}^n(\omega)+\beta\left[\hat{f}(\omega)-\sum^{K}\hat{u}_k^{n+1}(\omega)\right]$

**Step5**: Repeat(2)and(3), until $\frac{\sum_{k=1}^{K}\|\hat{u}_k^{n+1}-\hat{u}_k^n\|_2^2}{\|\hat{u}_k^n\|_2^2}<e$, the loop ends, and the modal components are obtained.

### *2.6 Execution process of the proposed hybrid model*

The proposed hybrid model is designed to enhance the accuracy of time-series forecasting, combining SSA, phase space reconstruction (PSR), Attention Gated Recurrent Unit networks (AtGRU) and VMD. The workflow is as follows:

**Step 1: Data Processing**

Using SSA, the original time-series data are broken down into their principal components, effectively denoising the data. This step removes noise and distills the data to its core dynamics, leveraging the embedding and singular value decomposition techniques described in equations (1) - (4).

**Step 2: Phase Space Reconstruction**

The denoised sequences are then reconstructed in phase space, a technique that captures the dynamics of the system and allows for more accurate modeling of complex, chaotic behavior. This involves determining the optimal embedding dimension and delay time using the mutual information and false nearest neighbor methods (equations (5) - (10)).

**Step 3: Preliminary Forecasting**

The At-GRU, an advanced variant of GRU networks incorporating attention mechanisms, processes the reconstructed phase space data for initial forecasting. The A-GRU uses the model described in equations (16) - (20) to account for both long-term dependencies and short-term fluctuations, while the attention module focuses the network on the most relevant information.

**Step 4: Error Forecasting**

The preliminary forecasts are then analyzed for errors, which are decomposed using VMD (2.5) into a series of Intrinsic Mode Functions (IMFs). This step, as per equations (21) - (26), is aimed at isolating and correcting the errors in the forecast, refining the predictive accuracy by focusing on the detailed error structure.

**Step 5: Final Forecasting**

The initial forecasting results and the error compensation sequences obtained from VMD are combined. This superimposition ensures that the final forecast considers both the initial predictive patterns and the refined error corrections, leading to a more accurate final outcome.

**Step 6: Model Validation**

Finally, the model's accuracy is validated using standard metrics such as Root Mean Square Error (RMSE), Mean Absolute Error (MAE), Mean Absolute Percentage Error (MAPE), and the Coefficient of Determination ($R^2$). The results are compared against the denoised sequence to evaluate the performance enhancements achieved through this hybrid modeling approach.

This execution process leverages the strengths of various advanced signal processing and machine learning techniques to deliver a robust and accurate time-series forecasting model, as visually summarized in **Fig.3.**

**Table 1** Characteristics of the gathered datasets

| Cases | Datasets | Numbers | Statistical indicators | | | |
|---|---|---|---|---|---|---|
| | | | Mean(m/s) | Std.(m/s) | Max.(m/s) | Min.(m/s) |
| | All Samples | 3251 | 8.3631 | 3.5523 | 19.9028 | 0.3115 |
| Site1 | Training Set | 2851 | 8.4972 | 3.5811 | 19.9028 | 0.3115 |
| | Testing Set | 400 | 7.4072 | 3.1795 | 14.4539 | 0.5560 |
| | All Samples | 2655 | 7.5884 | 2.2393 | 15.2390 | 0.4930 |
| Site2 | Training Set | 2255 | 7.3540 | 2.2353 | 15.2390 | 0.4930 |
| | Testing Set | 400 | 8.9104 | 1.7485 | 12.8263 | 4.3948 |
| | All Samples | 2950 | 7.1710 | 2.1628 | 14.4446 | 0 |
| Site3 | Training Set | 2550 | 7.1725 | 2.0608 | 14.4446 | 0 |
| | Testing Set | 400 | 7.1608 | 2.7250 | 13.2543 | 0.8140 |

**Table 2** Four evaluation metrics of all experiments

| Metric | Definition | Equation |
|---|---|---|
| **MAE** | The mean absolute error | $MAE = \frac{1}{N}\sum_{i=1}^{N}\lvert y_i - \hat{y}_i\rvert$ |
| **RMSE** | The square root of the average of error squares | $RMSE = \sqrt{\frac{1}{N}\sum_{i=1}^{N}(y_i - \hat{y}_i)^2}$ |
| **MAPE** | The average absolute percentage error | $MAPE = \frac{1}{N}\sum_{i=1}^{N}\left\lvert\frac{y_i - \hat{y}_i}{y_i}\right\rvert$ |
| **$R^2$** | R-square | $R^2 = 1 - \frac{\sum_{i=1}^{N}(y_i - \hat{y}_i)^2}{\sum_{i=1}^{N}(y_i - \bar{y})^2}$ |

**Note:** Other than $R^2$, the smaller the values, the better the prediction effect of the model.

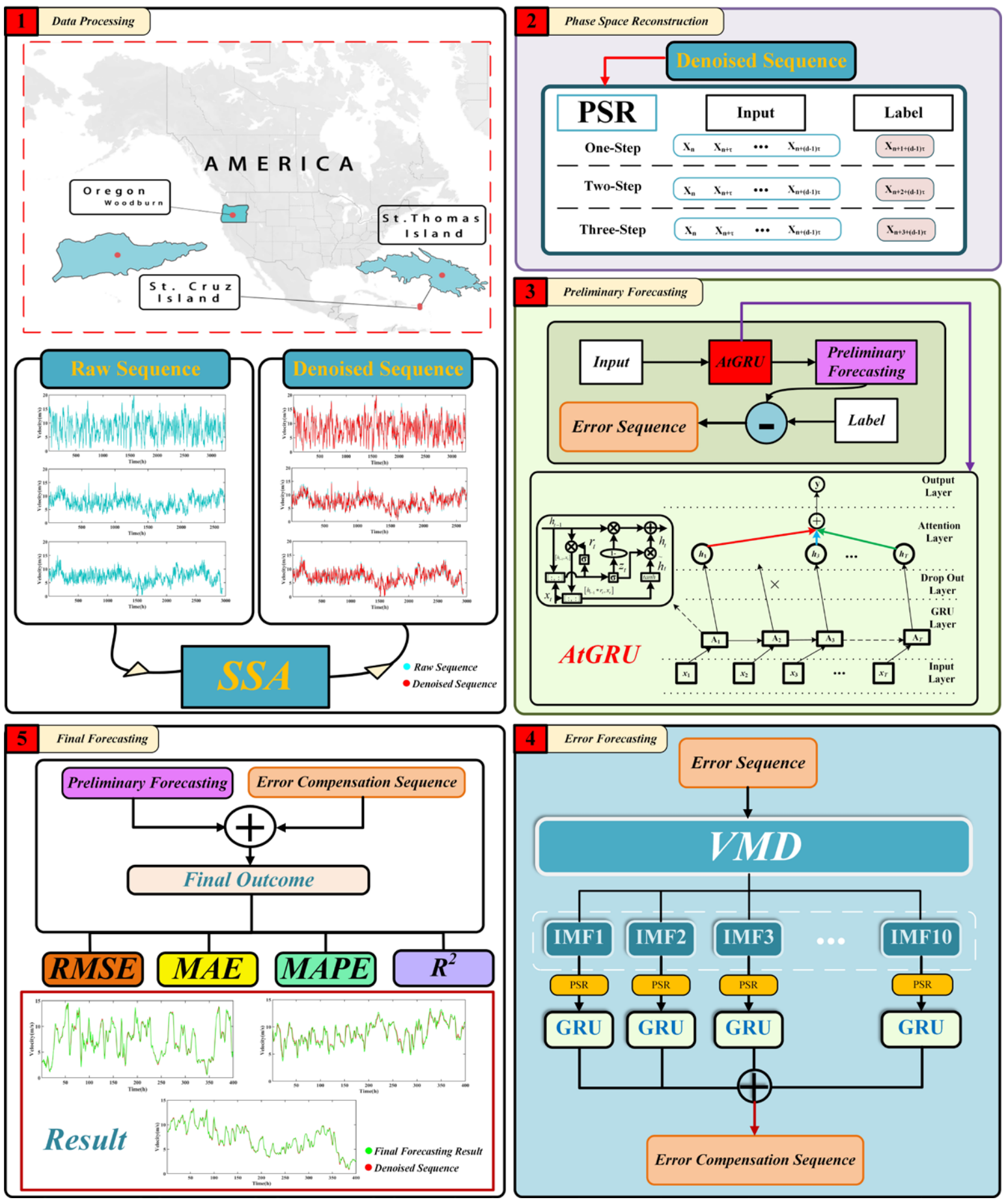

**Fig.4.**The flowchart of the proposed model

## 3 Experiment

### *3.1 Data description*

Inside this section, figures of 3 disparate wind farm sites of Woodburn, St. Thomas Island, and Santa Cruz Island with a measurement interval of 1h which are collected from January 1st, 2012to, December 31st, 2012, between January 1st, 2013 and January 1st, 2014 and between January 1st, 2013 and December 19th, 2013 are respectively used in the study. And their wind speed series correspond to the following three graphs, respectively. As shown in **Table 1**, the three datasets are divided into 400 samples as the test set. Their statistical characteristics are reflected in **Table 1**.

### *3.2 Parameter setting and evaluation metrics*

With the aim of accurately assessing the performance of the raised pattern, quantitative evaluation indicators are essential. **Table 2** shows the mean absolute error, square root of the error square average, an average of errors of the absolute percentage, and R-square were adopted for the purpose of assessing the test consequences. It ought to be noted that $N$ represents the length of the predicted sequence, and $y_i$ and $\hat{y}_i$ represent the true and predicted values, respectively.The model's parameters often exert an essential influence upon its training and prediction consequences. Suitable parameters were finally determined and shown in **Table 3**.

To ensure a precise assessment of raised pattern's performance, quantitative assessment metrics are necessary. The experimental results were evaluated using the average absolute error, the root average square error, average absolute percentage error, along with R-squared, as exhibited inside **Table 2**. It is important to note that $N$ denotes the length of the predicted sequence, while $y_i$ and $\hat{y}_i$ represent the predicted and actual values, separately. The selection of model parameters can significantly affect its training and prediction outcomes. After careful consideration, appropriate parameters were determined and presented in **Table 3**.

**Table 3** Model parameter setting

| Modules | Parameters | Determination methods | Values |
|---|---|---|---|
| PSR | Reconstruction dimension | Preset | 60 |
| | Time delay | Preset | 1 |
| EMD | No paramters | --- | --- |
| EEMD | noise_width | Preset | 0.05 |
| | trials | Preset | 100 |
| CEEMDAN | epsilon | Preset | 0.005 |
| | trials | Preset | 100 |
| | range_thr | Preset | 0.01 |
| | total_power_thr | Preset | 0.05 |
| | noise_scale | Preset | 1.0 |
| SSA | Embedding dimension | Preset | 20 |
| | Reconstruction dimension | Preset | 10 |
| VMD | Maximum number of iterations | Preset | 500 |
| | moderate bandwidth constraint | Preset | 5000 |
| | noise-tolerance(tau) | Preset | 0 |
| | DC | Preset | 0 |
| | the number of modes to be recovered | Preset | 10 |
| | init | Preset | 0 |
| | tolerance of convergence criterion | Preset | 1e-7 |
| SVR | kernel | Preset | 'linear' |
| BPNN | layers | Preset | 60/40/20/1 |
| | batch_size | Preset | 64 |
| | epochs | Preset | 150 |
| RNN | layers | Preset | 64 |
| | epochs | Preset | 150 |
| | batch_size | Preset | 64 |
| GRU | layers | Preset | 64 |
| | epochs | Preset | 150 |
| | batch_size | Preset | 64 |
| AtGRU | layers | Preset | 64 |
| | epochs | Preset | 150 |
| | batch_size | Preset | 64 |

**Table 5** Results of Experiment III and IV

| Site | Model | One-step | | | | Two-step | | | | Three-step | | | |
|---|---|---|---|---|---|---|---|---|---|---|---|---|---|
| | | RMSE | MAE | MAPE | $R^2$ | RMSE | MAE | MAPE | $R^2$ | RMSE | MAE | MAPE | $R^2$ |
| Site1 | SSA-AtGRU-CEEMDAN-GRU | 0.9962 | 0.1425 | 2.2889 | 0.1970 | 0.7620 | 0.5602 | 9.1480 | 0.9426 | 1.4763 | 1.1183 | 17.1878 | 0.7844 |
| | SSA-AtGRU-EMD-GRU | 0.2404 | 0.1761 | 2.8353 | 0.9943 | 0.8197 | 0.6184 | 10.2911 | 0.9335 | 1.4591 | 1.0902 | 17.1426 | 0.7894 |
| | SSA-AtGRU-EEMD-GRU | 0.2188 | 0.1421 | 2.3321 | 0.9953 | 0.7289 | 0.5322 | 8.6257 | 0.9475 | 1.2669 | 0.9270 | 14.6506 | 0.8412 |
| | SSA-AtGRU-SSA-GRU | 0.2341 | 0.1930 | 3.4288 | 0.9946 | 0.4848 | 0.3689 | 6.4644 | 0.9768 | 0.9020 | 0.6643 | 11.4221 | 0.9195 |
| | **SSA-AtGRU-VMD-GRU** | **0.1124** | **0.0844** | **1.4075** | **0.9988** | **0.3979** | **0.3061** | **4.9372** | **0.9843** | **0.7374** | **0.5382** | **9.4636** | **0.9462** |
| Site2 | SSA-AtGRU-CEEMDAN-GRU | 0.1293 | 0.1024 | 1.2354 | 0.9945 | 0.4012 | 0.3063 | 3.6769 | 0.9474 | 0.6793 | 0.5240 | 6.2809 | 0.8491 |
| | SSA-AtGRU-EMD-GRU | 0.1619 | 0.1307 | 1.5661 | 0.9940 | 0.4067 | 0.3098 | 3.7074 | 0.9459 | 0.6636 | 0.5241 | 6.3475 | 0.8559 |
| | SSA-AtGRU-EEMD-GRU | 0.0982 | 0.0746 | 0.9005 | 0.9968 | 0.3900 | 0.2994 | 3.6334 | 0.9503 | 0.5975 | 0.4582 | 5.6175 | 0.8832 |
| | SSA-AtGRU-SSA-GRU | 0.1138 | 0.0887 | 1.0589 | 0.9958 | 0.3088 | 0.2378 | 2.8357 | 0.9688 | 0.3679 | 0.2850 | 3.3259 | 0.9557 |
| | **SSA-AtGRU-VMD-GRU** | **0.0772** | **0.0607** | **0.7225** | **0.9980** | **0.2334** | **0.1828** | **2.1908** | **0.9822** | **0.3407** | **0.2651** | **3.2001** | **0.9620** |
| Site3 | SSA-AtGRU-CEEMDAN-GRU | 0.1120 | 0.0870 | 1.4327 | 0.9983 | 0.3904 | 0.2914 | 4.8360 | 0.9795 | 0.6960 | 0.5206 | 8.5679 | 0.9348 |
| | SSA-AtGRU-EMD-GRU | 0.1319 | 0.1013 | 1.6238 | 0.9977 | 0.4728 | 0.3774 | 6.5813 | 0.9700 | 0.6647 | 0.5016 | 7.9595 | 0.9405 |
| | SSA-AtGRU-EEMD-GRU | 0.1120 | 0.0892 | 1.5090 | 0.9983 | 0.3672 | 0.2645 | 4.4263 | 0.9818 | 0.6451 | 0.4943 | 8.1603 | 0.9440 |
| | SSA-AtGRU-SSA-GRU | 0.1474 | 0.1182 | 2.0422 | 0.9971 | 0.2830 | 0.2081 | 3.4820 | 0.9892 | 0.4036 | 0.3073 | 5.1441 | 0.9781 |
| | **SSA-AtGRU-VMD-GRU** | **0.0665** | **0.0499** | **0.8990** | **0.9994** | **0.2181** | **0.1683** | **2.7768** | **0.9936** | **0.3554** | **0.2775** | **4.9827** | **0.9830** |

**Table 4** Results of Experiment I and II

| Site | Model | One-step | | | | Two-step | | | | Three-step | | | |
|---|---|---|---|---|---|---|---|---|---|---|---|---|---|
| | | RMSE(m/s) | MAE(m/s) | MAPE(%) | $R^2$ | RMSE(m/s) | MAE(m/s) | MAPE(%) | $R^2$ | RMSE(m/s) | MAE(m/s) | MAPE(%) | $R^2$ |
| Site1 | SSA-SVR | 0.9638 | 0.7456 | 13.9730 | 0.9081 | 1.4715 | 1.1109 | 20.0641 | 0.7858 | 2.0776 | 1.5771 | 29.5530 | 0.5730 |
| | SSA-BPNN | 1.1509 | 0.8579 | 14.4978 | 0.8688 | 1.8453 | 1.3826 | 24.6509 | 0.6630 | 2.3513 | 1.7709 | 32.5789 | 0.4531 |
| | SSA-RNN | 0.3262 | 0.2435 | 4.0764 | 0.9894 | 1.0404 | 0.7943 | 13.5365 | 0.8926 | 2.0273 | 1.5564 | 27.5716 | 0.5922 |
| | SSA-GRU | 0.2717 | 0.2069 | 3.4812 | 0.9927 | 0.9431 | 0.7037 | 11.7571 | 0.9120 | 1.7449 | 1.3142 | 22.6267 | 0.6988 |
| | **SSA-AtGRU** | **0.2752** | **0.2062** | **3.4185** | **0.9925** | **0.9314** | **0.6950** | **11.6586** | **0.9142** | **1.7274** | **1.2813** | **21.7066** | **0.7048** |
| Site2 | SSA-SVR | 0.5799 | 0.4661 | 5.5744 | 0.8900 | 0.8153 | 0.6319 | 7.6598 | 0.7826 | 1.1217 | 0.8819 | 10.5312 | 0.5885 |
| | SSA-BPNN | 0.6276 | 0.4888 | 5.9252 | 0.8701 | 1.0246 | 0.8046 | 9.7939 | 0.6565 | 1.2298 | 0.9682 | 11.7345 | 0.5052 |
| | SSA-RNN | 0.2205 | 0.1696 | 2.0234 | 0.9838 | 0.6098 | 0.4751 | 5.7376 | 0.8777 | 1.2138 | 0.9542 | 11.2593 | 0.5175 |
| | SSA-GRU | 0.2048 | 0.1612 | 1.9023 | 0.9862 | 0.6004 | 0.4686 | 5.5659 | 0.8819 | 1.1050 | 0.8728 | 10.1431 | 0.5999 |
| | **SSA-AtGRU** | **0.1946** | **0.1538** | **1.8118** | **0.9875** | **0.5972** | **0.4625** | **5.5500** | **0.8831** | **1.0077** | **0.7883** | **9.4365** | **0.6678** |
| Site3 | SSA-SVR | 0.6644 | 0.5395 | 11.4138 | 0.9406 | 0.8669 | 0.7022 | 13.6821 | 0.8988 | 1.1984 | 0.9814 | 19.0320 | 0.8066 |
| | SSA-BPNN | 0.6073 | 0.4773 | 8.7496 | 0.9498 | 0.9958 | 0.8147 | 15.2100 | 0.8633 | 1.3266 | 1.0868 | 20.2200 | 0.7628 |
| | SSA-RNN | 0.1987 | 0.1564 | 2.6928 | 0.9946 | 0.6536 | 0.5127 | 9.1928 | 0.9423 | 1.2198 | 0.9758 | 18.6527 | 0.7992 |
| | SSA-GRU | 0.1807 | 0.1406 | 2.4150 | 0.9956 | 0.5492 | 0.4294 | 7.4904 | 0.9594 | 1.0453 | 0.8516 | 15.6877 | 0.8527 |
| | **SSA-AtGRU** | **0.1738** | **0.1348** | **2.4391** | **0.9959** | **0.5687** | **0.4486** | **7.7656** | **0.9563** | **1.0022** | **0.8025** | **14.9796** | **0.8647** |

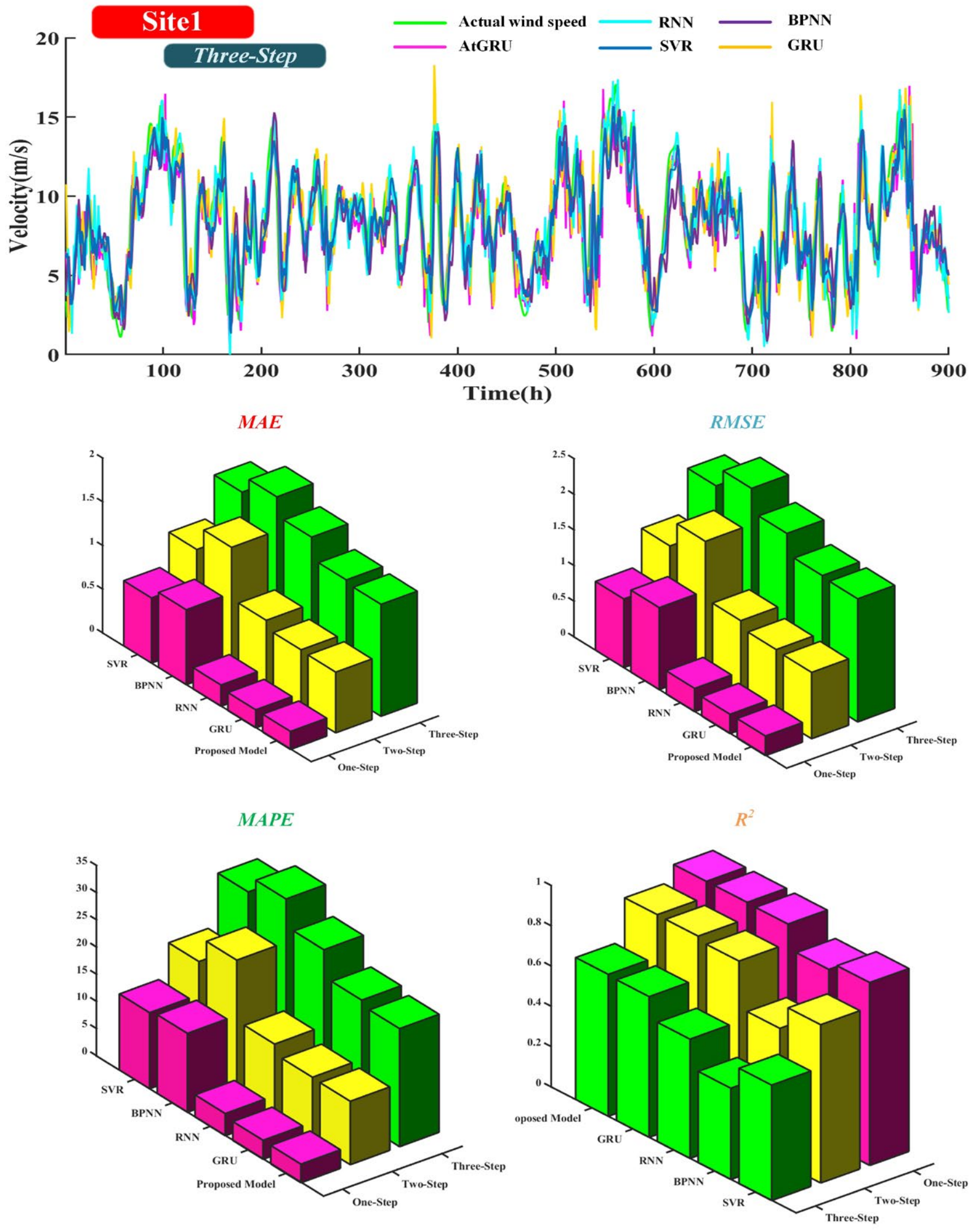

**Fig. 5.** Contrast of three-step forecast consequences in site1

### *3.3 Experiment I: Comparison of a single preliminary prediction model in the one-step forecast*

Experiment I proves that the raised pattern has greatly performance in one-step forecast than four original models, including SVR, BPNN, RNN, and GRU. As mentioned above, the sequences processed in Experiment I have all been processed by SSA noise reduction to eliminate the influence of measurement errors. As shown in Table 4, in the one-step forecast, the raised pattern obviously exceeds the SVR, BPNN, as well as RNN pattern in evaluating various indicators. For example, in Woodburn's measurement dataset site1, the raised pattern is significantly greater for the whole metrics in one-step prediction. Especially, in comparison to three patterns, the RMSE is reduced by 71.45%, 76.09% and 15.63%, the MAE is reduced by 72.34%, 75.96% and 15.32%, and the $R^2$ is improved by 9.29%, 14.24% and 0.31%, respectively. Since the proposed model parameters are nearly 67% larger than those of GRU, the proposed model requires large samples and more training iterations to be fully fitted, which sometimes makes it inferior to GRU. But when it fully fits the other two datasets, and it can outperform GRU. Therefore, in light of the consequences in **Fig.4**, it is capable of being concluded that the raised pattern own an elevated precision in one-step forecast.

### *3.4 Experiment II: Comparison of a single preliminary prediction model in the multi-step forecast*

Experiment II demonstrates that the raised pattern outperforms other four patterns in multi-step forecast. As could be known in **Table 4**, the raised pattern has the optimal performance. Regarding two-step forecast inside site1 as an instance, the proposed model reduces the RMSE by 36.70%, 49.53%, 10.48% and 1.24%, and 37.44%, 49.73%, 12.50%, respectively, than the SVR, BPNN, RNN and GRU, and improved $R^2$ by 14.05%, 37.89%, 2.42% and 0.24%, respectively. Still the best on site2 from St Thomas. The raised pattern performs the optimal in three-step forecast on all datasets. Still taking site1 as an example, the proposed model reduces RMSE by 55.17%, 60.39%, 54.06% and 1% , and MAE by 18.76%, 27.65%, 17.68% and 2.50% compared to SVR, BPNN, RNN and GRU, and improves by 23.00%, 55.55%, 19.01% and 0.86% $R^2$, respectively. Generally, when the proposed model is given enough training samples and training time, its prediction performance is the best on most datasets. Nevertheless, the quantity of parameters of the raised model exceeds 20,000, and thus does not exhibit ideal performance due to underfitting at some point. In **Table 4**, it can occasionally be observed that the raise pattern is slightly inferior to the forecast influence effect of the GRU pattern, verifying this shortcoming of the proposed pattern. However, in most scenarios, training sets with a sufficient sample size can often be provided for the training of model prediction, so the proposed model outperforms others in most scenarios.

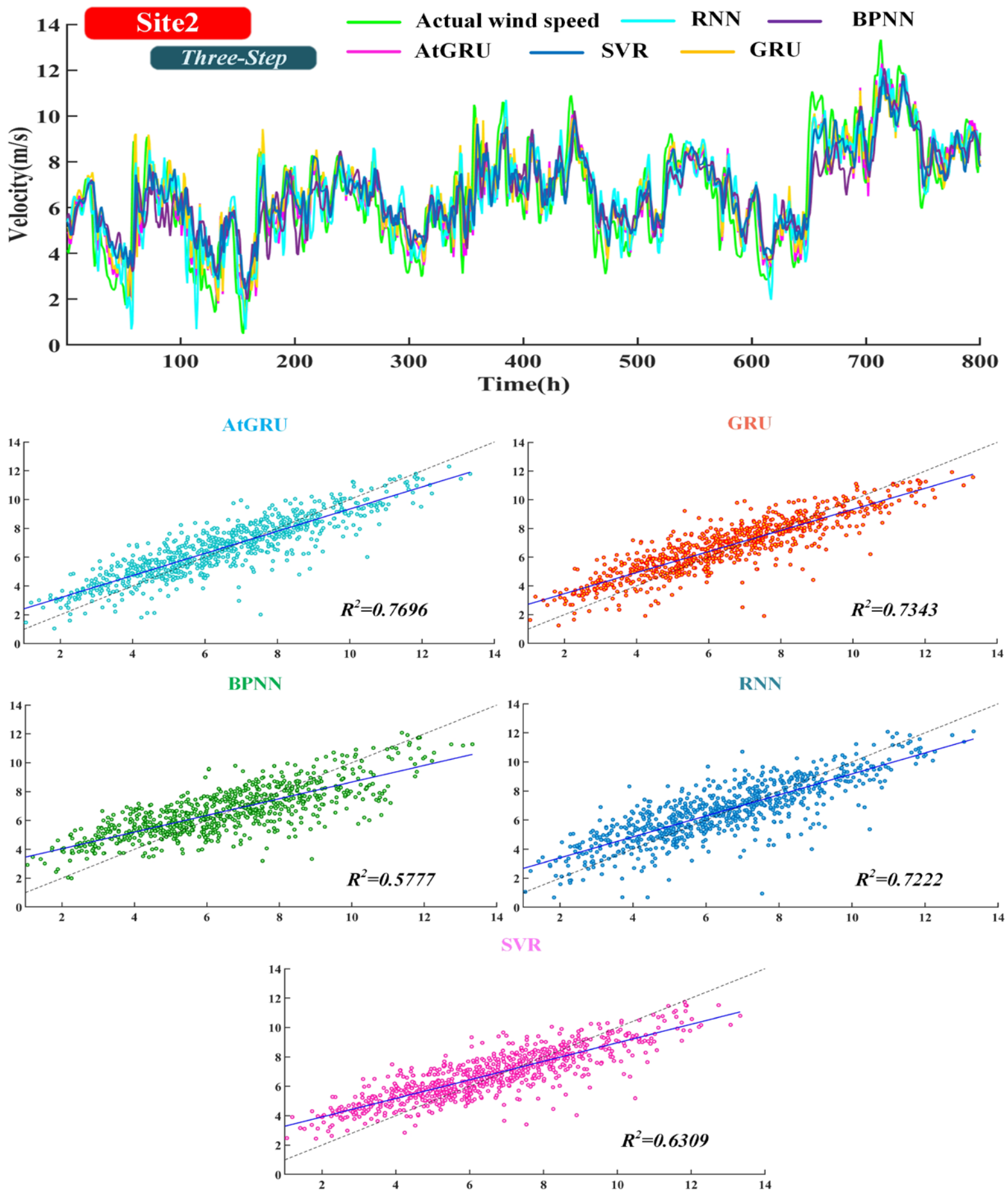

**Fig.6.**Contrast of three-step forecast consequences in site2

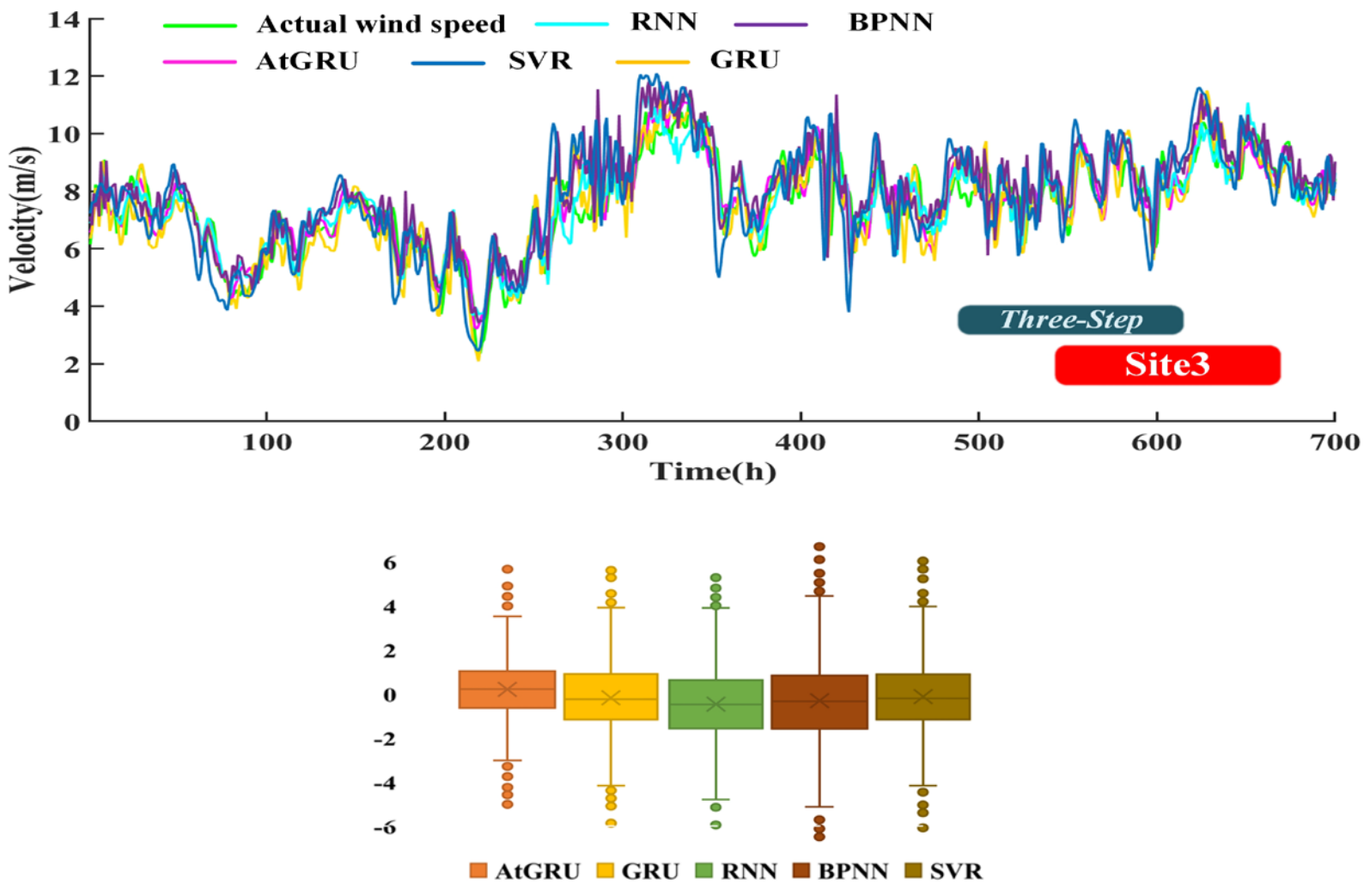

**Fig.7.** Contrast of three-step forecast consequences in site3

### *3.5 Experiment III: Comparison of sequence decomposition methods in the one-step forecast*

Experiment III proves that VMD has a better effect than other sequence decomposition methods in the one-step forecast. As shown in **Table 5**, this conclusion can be fully confirmed by combining the evaluation metrics after error correction. As shown in **Table 5**, in the one-step forecast, the error sequence processed by VMD is trained by GRU, and the final predicted error correction sequence is very consistent with the actual. Compared with **Table 4**, it can be figured out that the prediction effect after error correction becomes better at first. Next, it is able to be known from **Table 5** that in comparison to the other sequence decomposition methods, VMD has a more prominent role in promoting the effect of error correction. It is able to be known that in multi-step forecast, the processing VMD effect evidently greater than other methods, regardless of the number of prediction steps, which is closely related to the advantage that VMD is not easy to generate virtual modes in sequence decomposition. Due to the easy generation of virtual modes when decomposing sub-modes, EMD and its series of improved methods perform poorly, especially in three-step prediction. At the same time, it can be observed that the ability of SSA to extract the eigenmodes in the error sequence is not strong. Since the error values are highly randomly distributed, this feature may not be enough for SSA to obtain the main features conducive to deep learning models' training. It does not produce virtual modes, so its effect is second only to VMD in some cases, and the effect is the worst when some EMDs and their improvement methods are not prone to excessive virtual modes.

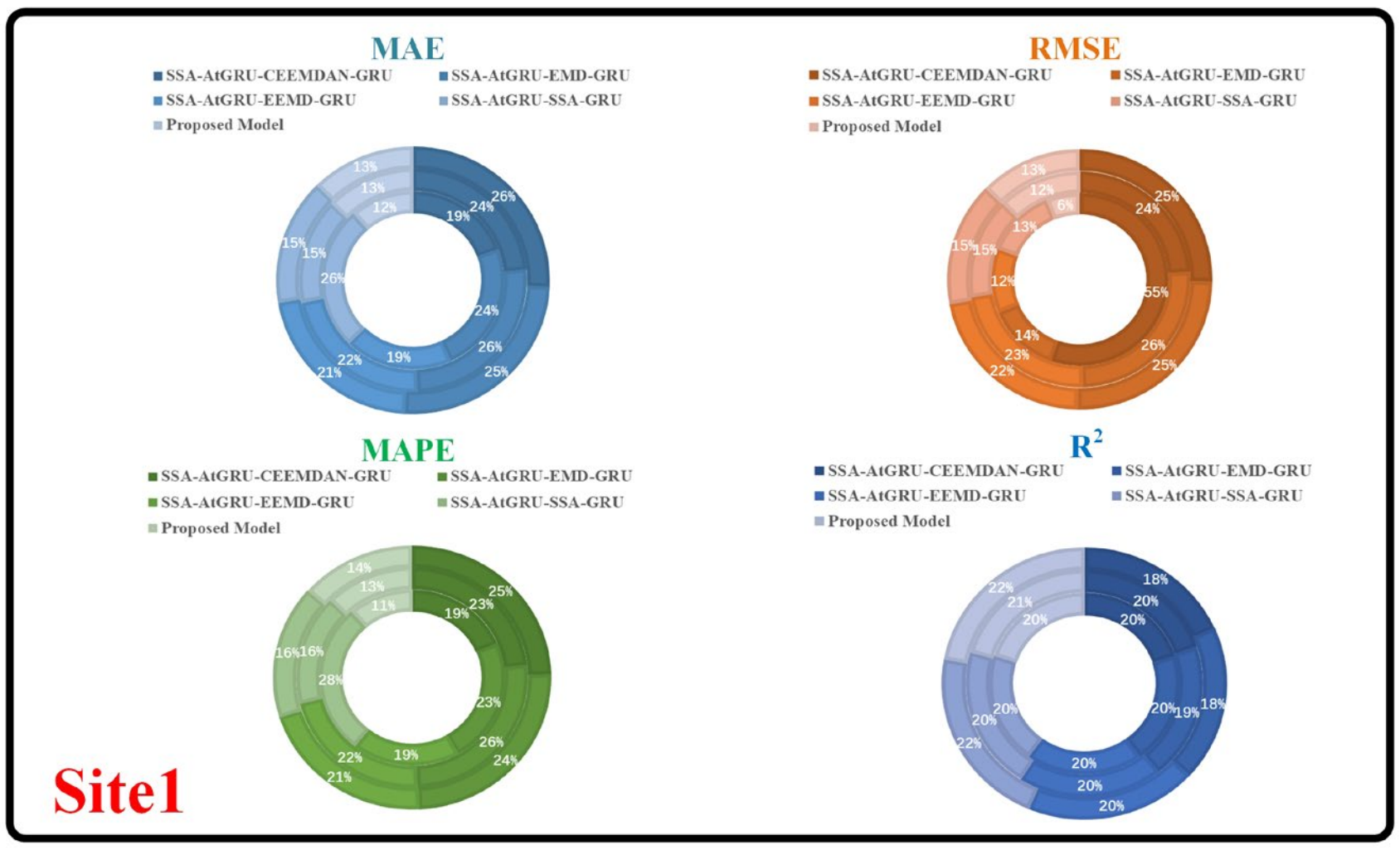

**Fig.8.** Contrast of decomposition methods in multi-step forecast consequences in site1

*3.6 Experiment IV: Comparison ofsequence decomposition methods in the multi-step forecast*

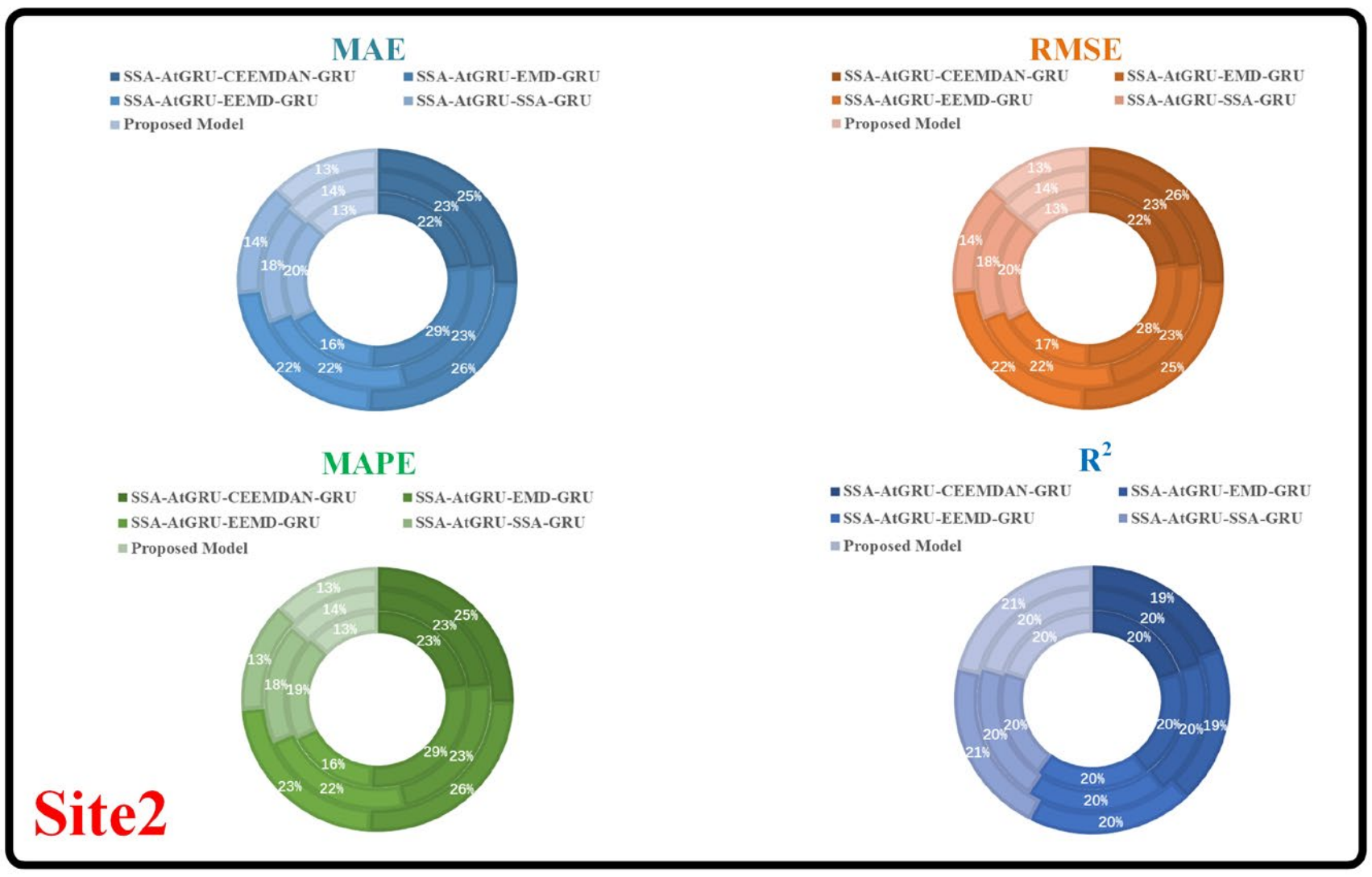

**Fig.9.** Contrast of decomposition methods in multi-step forecast consequences in site2

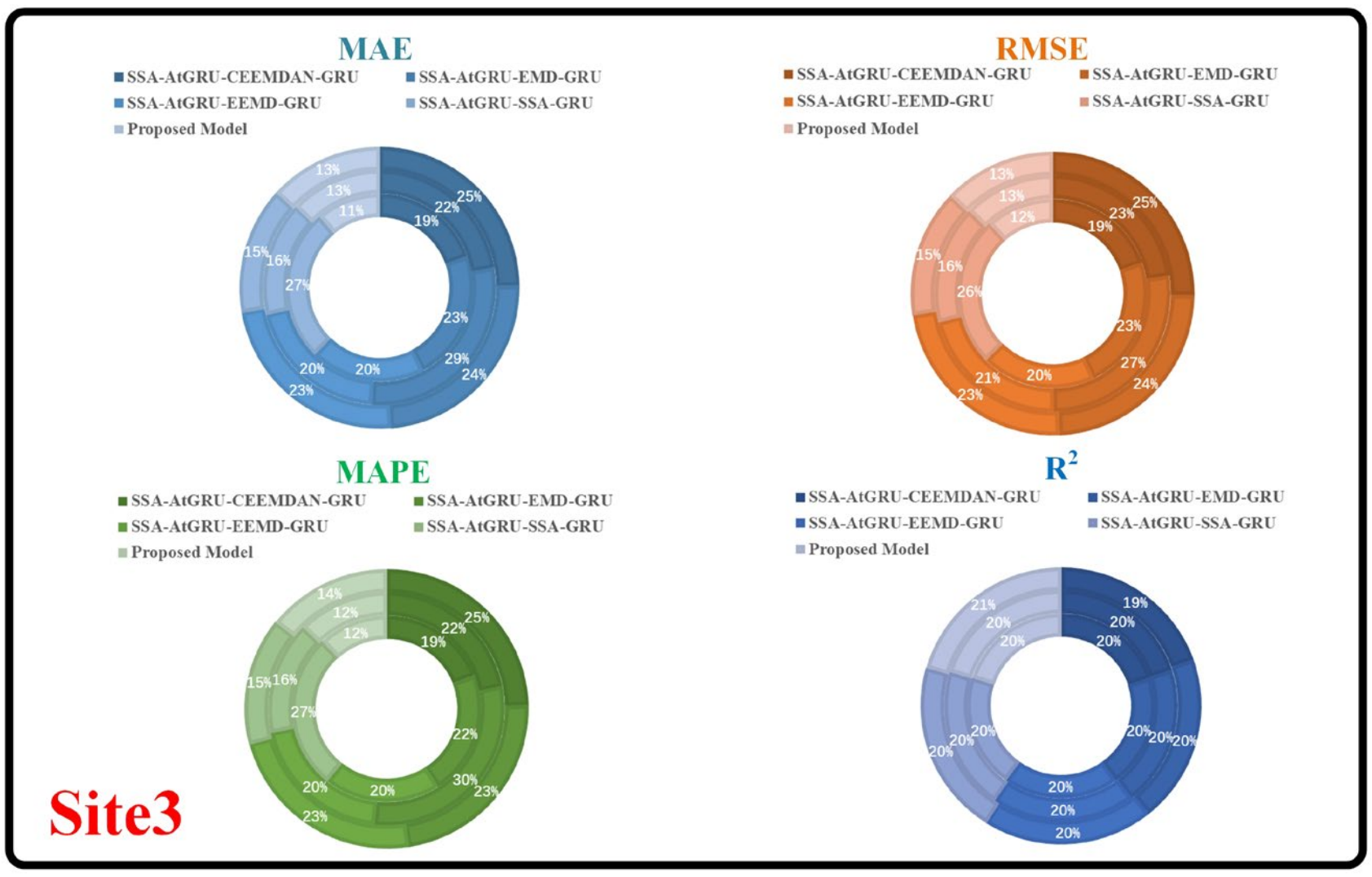

**Fig.10.** Contrast of decomposition methods in multi-step forecast consequences in site3

Experiment IV proves that the proposed model adopts the VMD method in the step of error sequence decomposition and has a better effect than other sequence decomposition approaches within multi-step forecast. As exhibited inside **Table 5**, it can be observed that the error sequences processed by VMD are all conducive to error correction. Comparing the results inside **Table 5**, it`s not hard to discover that the precision of the error correction consequences is improved considerably in the multi-step forecast. Synchronously, the processing effect of SSA is second only to that of VMD, which also coincides with the conclusion of the experimental analysis above for the comparison of sequence decomposition methods for a one-step forecast.

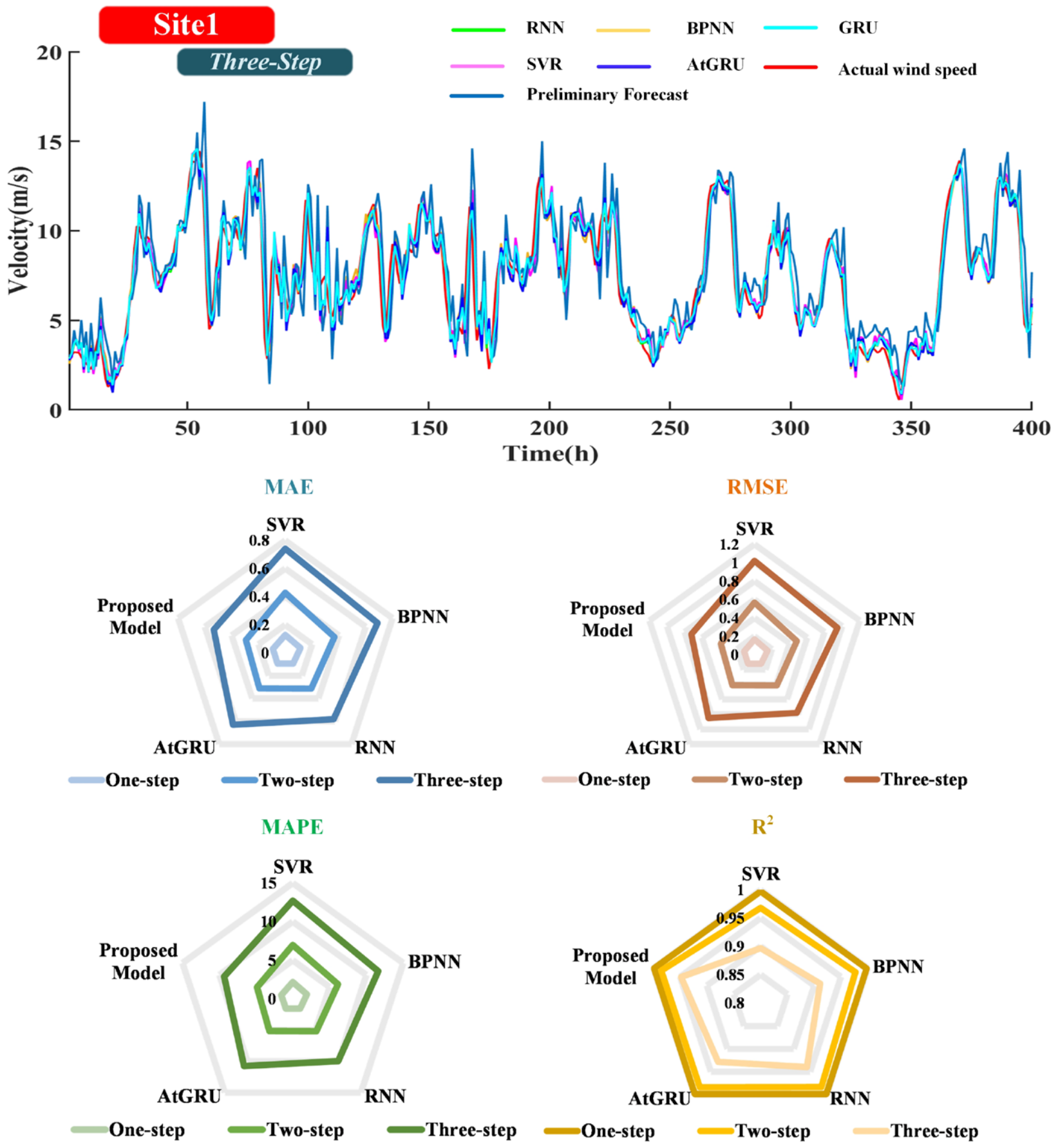

**Fig.11.** Comparison of forecast model based on error sequence in three-step forecast in site1

**Table 6** Results of Experiment V

| Site | Model | One-step | | | | Two-step | | | | Three-step | | | |
|---|---|---|---|---|---|---|---|---|---|---|---|---|---|
| | | RMSE | MAE | MAPE | $R^2$ | RMSE | MAE | MAPE | $R^2$ | RMSE | MAE | MAPE | $R^2$ |
| Site1 | SSA-AtGRU-VMD-SVR | 0.1667 | 0.1285 | 2.1628 | 0.9973 | 0.5673 | 0.4279 | 6.9696 | 0.9682 | 1.0206 | 0.7414 | 12.7230 | 0.8970 |
| | SSA-AtGRU-VMD-BPNN | 0.1469 | 0.1128 | 1.9113 | 0.9979 | 0.4811 | 0.3665 | 6.1278 | 0.9771 | 0.9466 | 0.6875 | 11.6210 | 0.9114 |
| | SSA-AtGRU-VMD-RNN | 0.1208 | 0.0935 | 1.5742 | 0.9986 | 0.4116 | 0.3131 | 5.1778 | 0.9832 | 0.7803 | 0.5824 | 9.9723 | 0.9398 |
| | SSA-AtGRU-VMD-AtGRU | 0.1208 | 0.0939 | 1.5652 | 0.9986 | 0.4098 | 0.3120 | 5.1512 | 0.9834 | 0.8506 | 0.6310 | 10.7632 | 0.9284 |
| | **SSA-AtGRU-VMD-GRU** | **0.1198** | **0.0929** | **1.5730** | **0.9986** | **0.3861** | **0.2945** | **4.8548** | **0.9853** | **0.7218** | **0.5341** | **9.3516** | **0.9485** |
| Site2 | SSA-AtGRU-VMD-SVR | 0.1005 | 0.0780 | 0.9268 | 0.9967 | 0.3135 | 0.2456 | 2.9649 | 0.9679 | 0.4545 | 0.3472 | 4.0913 | 0.9324 |
| | SSA-AtGRU-VMD-BPNN | 0.0877 | 0.0686 | 0.8241 | 0.9975 | 0.2728 | 0.2122 | 2.5495 | 0.9757 | 0.4169 | 0.3179 | 3.7988 | 0.9432 |
| | SSA-AtGRU-VMD-RNN | 0.0731 | 0.0571 | 0.6789 | 0.9983 | 0.2226 | 0.1733 | 2.0904 | 0.9838 | 0.3249 | 0.2542 | 3.0673 | 0.9655 |
| | SSA-AtGRU-VMD-AtGRU | 0.0730 | 0.0570 | 0.6758 | 0.9983 | 0.2279 | 0.1784 | 2.1524 | 0.9830 | 0.3523 | 0.2698 | 3.1973 | 0.9594 |
| | **SSA-AtGRU-VMD-GRU** | **0.0729** | **0.0571** | **0.6795** | **0.9983** | **0.2160** | **0.1685** | **2.0348** | **0.9847** | **0.3128** | **0.2406** | **2.8831** | **0.9680** |
| Site3 | SSA-AtGRU-VMD-SVR | 0.0987 | 0.0765 | 1.3205 | 0.9987 | 0.3039 | 0.2350 | 4.1511 | 0.9876 | 0.5183 | 0.4023 | 7.0783 | 0.9638 |
| | SSA-AtGRU-VMD-BPNN | 0.0876 | 0.0667 | 1.1397 | 0.9990 | 0.2759 | 0.2080 | 3.5383 | 0.9898 | 0.4923 | 0.3767 | 6.5338 | 0.9674 |
| | SSA-AtGRU-VMD-RNN | 0.0685 | 0.0531 | 0.9142 | 0.9994 | 0.2080 | 0.1613 | 2.7522 | 0.9942 | 0.3850 | 0.3000 | 5.2642 | 0.9800 |
| | SSA-AtGRU-VMD-AtGRU | 0.0694 | 0.0539 | 0.9257 | 0.9994 | 0.2123 | 0.1652 | 2.8893 | 0.9939 | 0.3993 | 0.3122 | 5.3414 | 0.9785 |
| | **SSA-AtGRU-VMD-GRU** | **0.0695** | **0.0540** | **0.9279** | **0.9994** | **0.2041** | **0.1590** | **2.7017** | **0.9944** | **0.3505** | **0.2726** | **4.6381** | **0.9835** |

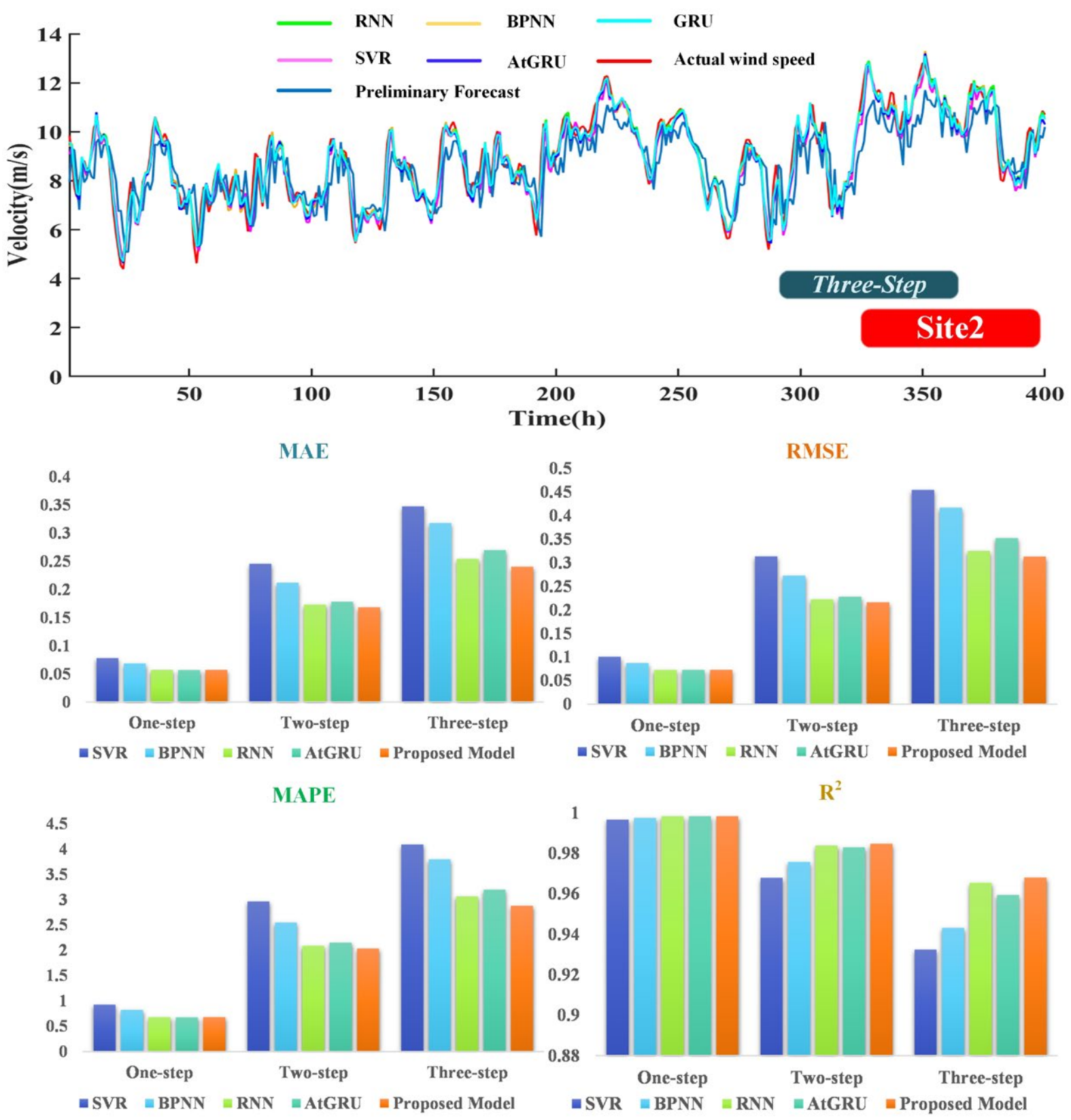

**Fig.12.** Comparison of forecast model based on error sequence in three-step forecast in site2

### *3.7 Experiment V: Comparison of a prediction model based on error sequence*

In the error sequence prediction scenario, due to the highly random distribution of errors, the convergence rate of the model is significantly slowed down during the training process. If a model with a huge number of parameters gets used for training, likely, ideal prediction results can not be obtained. Obviously, the prediction performance is straightforward associated with the quantity of the parameters of the model under limited training samples and limited training duration. Therefore, GRU can be well qualified for the error correction. As shown in **Table 6**, GRU has the best prediction on error sequences in most datasets for the same error sequence. Therefore, it`s easy to infer that the raised pattern performs best in most scenarios.

## 4 Discussions

Inside this part, benefits of the raised model is going to be disputed. In the above results, we can get some points as follows:

***I. Uniqueness***

The model raised inside this article consists of a predictor and an error corrector based on the error correction strategy. The predictor uses a combination of GRU model and attention to enhance the capability of extracting beneficial characteristics from the complex mixed modalities in the historical wind speed series, resulting in a high accuracy predictor. The VMD is adopted in a bid to divide the error sequence, and the GRU model is adopted with a view to efficiently extracting features from the error sequence within a limited amount of training samples and time, which leads to the construction of an excellent error corrector.

***II. Easy to operate and low training cost***

Compared with the preceding ones, the combination pattern proposed in this essay greatly simplifies the workflow. Concretely speaking, on the one hand, the proposed model does not use a series of optimization algorithms to optimize various parameters related to model performance but adopts an error correction strategy. This strategy greatly simplifies the workflow, making the model easier to operate and achieving high prediction accuracy. On the other hand, the model could have set a large number of parameters in error correction to have one neural network model with better prediction performance. However, considering that the prediction error always presents a highly random distribution in practical application scenarios, under the condition of limited training samples and training time, using a model of the neural network blessed with many parameters with the objective to predicting the erroneous sequence is usually ineffective. Consequently, the proposed model adopts the GRU model with a small number of parameters, making it easier to converge and predicting the error sequence with a good prediction effect. The test consequences indicate that this approach not merely reduces the training time, but decreases the number of samples needed to improve model performance.

***III. The proposed model has good robustness.***

The proposed model excludes the influence of irrelevant variables to obtain the consequences in this research. In addition to the experimental results presented in this paper, sensitivity tests are also employed in the raised pattern. By changing the parameters exhibited inside **Table 3** and using the raised pattern to other nonlinear non-stationary sequences, the test consequences indicate that the model presented in this article has a great steadiness in the multi-step forecasts in multiple scenarios.

The pattern raised in this research eliminates the impact of irrelevant variables to ensure accurate results. Sensitivity tests were conducted to assess the proposed pattern in addition to the test consequences presented in this paper. By adjusting the parameters listed in **Table 3** and applying the proposed model to other nonlinear non-stationary sequences, the test consequences expound that the pattern exhibits excellent stability and high performance in multi-step forecasting across various scenarios.

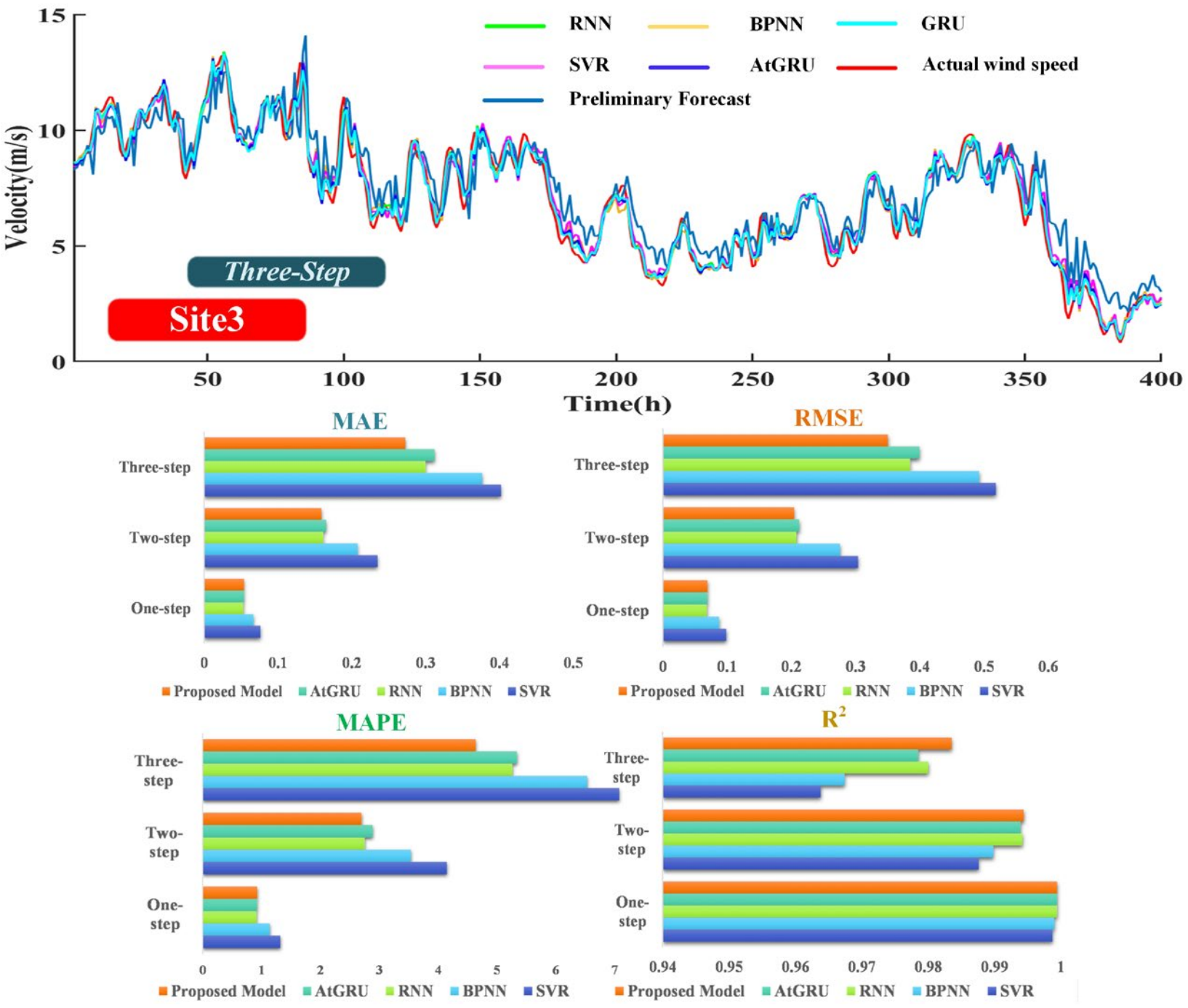

**Fig.13.** Comparison of forecast model based on error sequence in three-step forecast in site3

## 5 Conclusions

The energy of wind is one category of clean, renewable and abundant energy, playing an important role of green energy up to now. High-precision prediction of its speed conduces to ensuring the efficiency and security of the energy grid dispatching. However, wind speed series are often non-stationary and nonlinear, bringing great challenges to forecast. At present, many related solutions are proposed on the foundation of the physical approaches, statistical approaches, and agent combination methods, with few studies based on error correction frameworks. This article boosts a theoretical framework of error correction based on the research on a large number of solutions proposed by predecessors. It raises a blended short-run wind speed prediction pattern on the foundation of an attention-gated recurrent neural network and strategy of error correction. Specifically, the pattern is divided into two sub-models: the AtGRU predictor and the GRU error corrector. Firstly, historical wind speed series is filtrated by SSA noise reduction with the aim of filtering out the noisy part of wind speed measurement, and the noise sequence is reconstructed into multi-dimensional wind speed series by PSR, which lays the stage for subsequent multi-step model prediction. Next, the predictor is instructed on the foundation of historical ones, and corresponding mistake sequence is achieved for training the error corrector. In this regard, the error sequence is firstly passed through the VMD and divided into an array of the sub-modal sequences, serving as the training input of the GRU error corrector. Thus, the error corrector gets trained well and will make predictions about the error. Finally, through superimposing the preliminary forecast consequences of the predictor and error correction results predicted by the corrector one by one, and error-corrected wind speed prediction sequence is obtained. In terms of the experimental analysis above, the conclusions below are able to be drawn:

$(\boldsymbol{i})$.In the preliminary forecast stage, the AtGRU model can obtain high-precision wind speed prediction in both one-step forecast and multi-step forecast and obtain a relatively simple error sequence, which strongly supports the simplification of error correction work, especially in the multi-step forecast. From the consequences of Experiment I and Experiment II, AtGRU in the three-step forecast on the foundation of the three datasets used in this paper ($site1$~$site3$) is reduced by 4.07%, 6.97%, and 4.51%, respectively, compared with the ordinary GRU model without attention on MAPE.

$(\boldsymbol{ii})$.For the error sequence obtained by preliminary prediction, VMD is adopted with the objective to decomposing the sequence of errors. Besides, the features of the error sequence can be extracted in multi-step and single-step forecasts. Finally,the high-precision prediction can be obtained in the final error correction. From the results of Experiment III and Experiment IV, taking site1 as an example, the error correction results processed by VMD are CEEMDAN, EEMD, EMD, and SSA, respectively decreased by 44.94%, 44.79%, 35.04%, 17.15% on MAPE.

$(\boldsymbol{iii})$.For the training of error sequences, $GRU$ works as the predictor, which can obtain high-precision prediction in multi-step and one-step forecasts at a lower training cost. Experiment V shows that in the three-step forecast based on $site1$, the proposed model is reduced by 26.50%, 19.53%, 6.22%, 13.12%, on MAPE, respectively, and reduced MAPE by 56.92% concerning preliminary prediction.

Consequently, the pattern could effectively ameliorate the forecast precision of the wind speed series with limited training samples and training time For the forecast scenarios of nonlinear non-stationary sequences, the combined forecast pattern raised in this essay could not just offer the precise wind speed information for the efficiency as well as safety of wind power dispatching but also effectively assist decision-making in lots of domains, including stock marketplace, traffic management and so on.

Funding Statement
This research accepted not any external funding
Interest Conflicts
The writers doesn`t claim any interest conflict.
Information Availability Statement
The information which back up the discoveries of this research are accessible in this paper.